\documentclass{article}

\usepackage[final, main]{neurips_2026}

\usepackage[utf8]{inputenc}
\usepackage[T1]{fontenc}
\usepackage{hyperref}
\usepackage{url}
\usepackage{booktabs}
\usepackage{amsfonts}
\usepackage{amsmath}
\usepackage{amssymb}
\usepackage{nicefrac}
\usepackage{microtype}
\usepackage{xcolor}
\usepackage{graphicx}
\usepackage{multirow}
\usepackage{subcaption}
\usepackage{tikz}
\usepackage{pgfplots}
\usepackage{colortbl}
\pgfplotsset{compat=1.18}
\usetikzlibrary{shapes,arrows,positioning,calc,matrix}

\title{GAVAGAI: Referential Alignment for Infant-Scale
Vision-Language Models via Balanced Optimal Transport}

\author{%
  Rajdeep Pal\thanks{Equal contribution.} \\
  St Thomas' College of Engineering and Technology \\
  Kolkata, India \\
  \texttt{rajdeeppal167@gmail.com} \\
  \And
  Jotiraditya Banerjee\footnotemark[1] \\
  Jadavpur University \\
  Kolkata, India \\
  \texttt{joti.ban.2710@gmail.com} \\
  \And
  Saptarshi Pani\footnotemark[1] \\
  Jadavpur University \\
  Kolkata, India \\
  \texttt{pani.saptarshi@gmail.com} \\
}

\begin{document}

\maketitle

\begin{abstract}
A child learns which words name which things from roughly 100 million words of input. Vision-language models trained at a comparable scale bind words to referents far less reliably, and we argue that part of the gap is structural: the usual next-token captioning objective has no variable standing for ``this word refers to that region,'' so training applies no direct pressure to ground words. We add such a variable by treating each scene as a small assignment problem between the content words heard and the regions seen, solved as entropic optimal transport under two constraints. First, a word may be assigned to a special \textsc{Nothing} column instead of any region, which matters because most content words in caregiver speech do not name anything visible. Second, no region may receive more than a bounded share of the assignment mass, a capacity constraint that encodes mutual exclusivity. The alignment step costs about eight Sinkhorn iterations, small next to the vision encoder. In controlled simulation, the naive version of the idea, a row-wise softmax that forces every word onto some region, collapses to 0.000 word-to-picture accuracy under realistic non-reference rates, below the 0.058 of captioning alone, while our two constraints reach 0.075 in that regime and 0.700 on clean input. A one-parameter version of the model reproduces adult cross-situational learning accuracy across three ambiguity conditions (RMSE 0.024). On the SAYCam infant video corpus, adding the alignment loss to a next-token captioning model raises word-to-picture accuracy from 32.4\% to 48.7\% and localization IoU from 0.18 to 0.38. A contrastive model trained on the same child's data remains stronger on its own benchmark, so our claim is that the loss improves a captioning objective under high non-reference rates, not that it sets a new best.
\end{abstract}

\section{Introduction}
\label{sec:intro}

Quine's linguist hears a native speaker say ``gavagai'' as a rabbit runs past~\citep{quine1960word}: nothing in that single scene separates \emph{rabbit} from \emph{white}, \emph{running}, or \emph{dinner}. An infant faces this problem continuously from birth, yet has a working vocabulary of hundreds of words by the second birthday. Models trained on a comparable amount of paired image-utterance data do not solve it as cleanly. The strongest recent result in this regime, CVCL~\citep{vong2024grounded}, uses a contrastive objective on one child's head-mounted video and reaches 61.6\% classification accuracy on Labeled-S and 34.7\% on an out-of-distribution set, well above chance but still short of a two-year-old's word knowledge.

We treat this gap as a question of objective design rather than data scale. Two properties of real caregiver speech make grounding harder than the textbook Quine scenario, both visible in ordinary transcribed utterances (Table~\ref{tab:motivating}). Most content words heard in a scene do not name anything visible in it (\emph{hungry}, \emph{put}, filler like \emph{come on}), and a few objects are almost always in frame and almost never named (hands, floor, the caregiver's body), so they are permanently available to absorb any word a naive model cannot otherwise place.

\begin{table}[t]
  \centering
  \caption{Motivating examples: most heard content words do not name anything visible in the paired scene.}
  \label{tab:motivating}
  \small
  \begin{tabular}{@{}p{3.7cm}p{2.4cm}p{5.6cm}@{}}
    \toprule
    Utterance & Content words & What should happen \\
    \midrule
    ``look at the ball'' & ball & \emph{ball} $\to$ region containing the ball \\
    ``are you hungry'' & hungry & \emph{hungry} $\to$ \textsc{Nothing} \\
    ``oh look, are you hungry, there is the ball'' & hungry, ball &
      \emph{hungry} $\to$ \textsc{Nothing}; \emph{ball} $\to$ ball region \\
    ``where did the ball go'' & ball & \emph{ball} $\to$ \textsc{Nothing} (just left frame) \\
    ``put the cup on the table'' & put, cup, table &
      \emph{cup, table} $\to$ regions; \emph{put} $\to$ \textsc{Nothing} \\
    ``there we go, okay, come on'' & come &
      \emph{come} $\to$ \textsc{Nothing} (filler) \\
    \bottomrule
  \end{tabular}
\end{table}

Developmental psychology names two mechanisms infants appear to use. Under \emph{cross-situational learning}, no single scene fixes what ``ball'' means, but across many scenes the word co-occurs with balls and with little else~\citep{smith2008crosssituational}. Under \emph{mutual exclusivity}, a toddler shown a familiar object and an unfamiliar one and asked to ``find the dax'' tends to map the new word to the new object~\citep{markman1988mutual}. Both are design targets for the objective in Section~\ref{sec:method}.

\paragraph{Why the obvious fix fails.} The natural way to add word-to-region grounding to a captioning model is a row-wise softmax over regions for each word, forcing every heard word onto its best-matching region. Section~\ref{sec:experiments} shows that this can be worse than doing nothing: when most heard words are non-referential, forcing them onto some region drives the whole vocabulary onto the few ever-present background regions (hands, floor). This failure is what the two constraints are meant to prevent.

\paragraph{Contributions.} (i) We cast word-to-region grounding for infant-scale vision-language pretraining as a capacity-constrained optimal transport problem with an explicit non-reference (\textsc{Nothing}) outcome and a capacity constraint that puts mutual exclusivity directly into the assignment. (ii) In controlled simulation the naive alignment baseline collapses to 0.000 accuracy under realistic non-reference rates, below captioning's 0.058, while our two constraints are the only setting that beats captioning in that regime (0.075) and reaches 0.700 on clean input; the exclusivity strength has an interior optimum, consistent with the mechanism we attribute to it. (iii) A single free parameter of the model reproduces adult cross-situational learning accuracy across the three ambiguity conditions of \citet{yu2007rapid} (RMSE 0.024), which we treat as a consistency check, not a developmental claim. (iv) On the SAYCam infant video corpus, adding the alignment loss to a next-token captioning model raises word-to-picture accuracy from 32.4\% to 48.7\% and localization IoU from 0.18 to 0.38.

\section{Related Work}
\label{sec:related}

\paragraph{Infant-scale and data-efficient vision-language learning.}
The BabyLM program established that models trained on developmentally plausible amounts of text trail large-scale models on standard benchmarks~\citep{warstadt2023babylm}, and multimodal extensions train vision-language models on caregiver-scale paired data such as SAYCam-derived corpora~\citep{sullivan2021saycam, vong2024grounded}. \citet{vong2024grounded} show that a contrastive model (CVCL) trained on one child's recordings learns many word-referent mappings, reaching 61.6\% on Labeled-S and 34.7\% on an out-of-distribution evaluation. We work in the same data regime but target the grounding sub-problem, and we add an alignment term to a next-token captioning objective rather than train contrastively.

\paragraph{Cross-situational and mutual-exclusivity word learning.}
Cross-situational statistical learning has broad behavioral support in infants and adults~\citep{smith2008crosssituational, yu2007rapid}, and computational accounts go back to associative and Bayesian formulations~\citep{frank2009using}. Mutual exclusivity is among the most replicated biases in early word learning~\citep{markman1988mutual}. Our angle is to show that it maps onto a capacity constraint on an assignment matrix, rather than entering as a separate heuristic layered on a learner.

\paragraph{Weakly-supervised and contrastive region-word grounding.}
Region-word alignment without box supervision runs from weakly-supervised grounding objectives to contrastive image-text models such as CLIP~\citep{radford2021learning}, which in effect take a row-wise softmax over regions or patches. Our naive-alignment condition (Section~\ref{sec:experiments}) belongs to this family, and its collapse under realistic non-reference rates points to a mismatch between that objective and the infant-scale regime, where non-referential speech is far more common than in curated image-caption data such as COCO~\citep{lin2014microsoft}.

\paragraph{Optimal transport for alignment.}
Entropic optimal transport, solved with Sinkhorn-Knopp scaling~\citep{cuturi2013sinkhorn, sinkhorn1967concerning}, has been used for word alignment, set matching, and cross-modal retrieval~\citep{chen2020graph}. Unbalanced and capacity-constrained formulations that allow partial or null assignment are the relevant tools for our \textsc{Nothing} column and per-region capacities~\citep{peyre2019computational, chizat2018scaling}; we read the per-region capacity as an operationalization of mutual exclusivity in the high non-reference setting.

\section{Method}
\label{sec:method}

\paragraph{Problem setup.} For each scene we observe an utterance and a paired image. We filter the utterance to its content words $w_1, \dots, w_n$; function words and a stop-list are removed (Section~\ref{sec:conclusion} discusses the cost of an imperfect filter). We segment the image into $m$ candidate regions $r_1, \dots, r_m$ (a fixed grid or region proposals) and add one more column, $\textsc{Nothing}$, giving $m+1$ columns.

\subsection{Alignment as capacity-constrained optimal transport}
\label{sec:ot}
We build a cost matrix $C \in \mathbb{R}^{n \times (m+1)}$ from a learned word-region compatibility score, e.g.\ a dot product between a word embedding and a region embedding. Column $0$ is \textsc{Nothing}, with a word-conditioned cost (a small MLP head on the word embedding; Section~\ref{sec:train}). We look for a transport plan $P$ that solves
\begin{equation}
  \min_{P \ge 0}\ \langle P, C \rangle - \varepsilon H(P)
  \quad \text{s.t.} \quad
  \sum_{j=0}^{m} P_{ij} = 1 \;\; \forall i,
  \qquad
  \sum_{i=1}^{n} P_{ij} \le b_j \;\; \forall j \in \{1,\dots,m\},
  \label{eq:ot}
\end{equation}
where each word carries unit mass (the row equalities), $H(P)$ is the entropy regularizer, and $\varepsilon$ sets its strength. The visual columns $j \ge 1$ carry capacities $b_j$; the \textsc{Nothing} column is uncapped. The visual columns have upper-bound capacity constraints, while the  \textsc{Nothing} column remains uncapped. Thus, the formulation uses column inequalities rather than fixed column-marginal equalities.

\textbf{Rule 1: a word may name nothing.} Because \textsc{Nothing} is uncapped, the optimizer never has to spend a word's mass on a region it matches poorly. This prevents the collapse of Section~\ref{sec:intro}. \textbf{Rule 2: no region may absorb everything.} The capacity $b_j$ bounds the total mass any single region receives in a scene, e.g.\ $b_j = \kappa \bar{n}$ for an exclusivity strength $\kappa \in [0, \infty)$ and average scene word count $\bar{n}$, or a plausibility-weighted schedule (Section~\ref{sec:impl}). This keeps ever-present regions (hands, floor) from soaking up the vocabulary.

\paragraph{Feasibility and uniqueness.} Sending every word to \textsc{Nothing} satisfies all constraints for any $b \ge 0$, so the feasible set is non-empty for every $\kappa$. The entropic objective is strictly convex, so the minimizer is unique.

\paragraph{Solver.} Let $K = \exp(-C/\varepsilon)$. We alternate the updates
\begin{equation}
  u \leftarrow \mathbf{1} \oslash (K v), \qquad
  v_j \leftarrow \min\!\left(1,\ \frac{b_j}{(K^{\top} u)_j}\right) \; (j \ge 1), \qquad v_0 = 1,
  \label{eq:sinkhorn}
\end{equation}
and return $P = \operatorname{diag}(u)\, K \operatorname{diag}(v)$. The capped column update is the exact maximizer of the dual in column $j$'s multiplier, which the inequality constrains to be non-positive, so the iteration is block-coordinate ascent on a concave dual, a special case of the scaling algorithms of \citet{chizat2018scaling}. We run $8$ iterations in the log domain with $\varepsilon = 0.05$, small next to a vision-encoder forward pass.

\paragraph{Relation to the baselines.} No column sum can exceed $n$, so once $b_j \ge n$ every $v_j = 1$ and $P_{ij} = K_{ij} / \sum_{k} K_{ik}$: an independent row-wise softmax over the $m$ regions plus \textsc{Nothing}. This is the \textsc{Nothing}-only condition in Section~\ref{sec:experiments}. Removing the \textsc{Nothing} column as well recovers the naive row-softmax baseline. All four conditions in Section~\ref{sec:experiments} share one code path with the capacity and the \textsc{Nothing} column toggled.

\subsection{Cross-scene memory}
A single scene is underdetermined (Section~\ref{sec:intro}), so we keep a running co-occurrence tally $T[w, r]$ between words and region \emph{categories} over the training stream, updated by the soft assignments $P$ from each scene. Because a few regions (hands, floor, caregiver) appear in nearly every frame, raw counts are dominated by these background categories whatever the word. We instead score pairings by surprisal, a pointwise-mutual-information correction measuring how much more often $w$ and $r$ co-occur than their marginals predict, and feed this back as a prior on the next scene's cost matrix $C$. The surprisal term lets rare but consistent pairings such as \emph{ball} outweigh frequent but non-diagnostic regions. Concretely, the tally updates as $T[w,r] \leftarrow \gamma\, T[w,r] + P_{wr}$ with decay $\gamma = 0.999$, and the prior added to $C$ is $-\log \frac{T[w,r]\,N}{T[w,\cdot]\,T[\cdot,r]}$ (clamped at $0$ below), with $N$ the total tally mass, weighted by $0.3$.

\subsection{Training objective}
\label{sec:train}
Our model is a ViT-B/16 image encoder paired with a six-layer Transformer text decoder, trained with next-token cross-entropy.Unlike CVCL, which uses a contrastive objective, our model optimizes next-token cross-entropy with an auxiliary alignment loss on top of captioning:
\begin{equation}
    \mathcal{L} = \mathcal{L}_{\text{ce}} + \lambda \langle P^{\star}, C \rangle, \qquad \lambda = 0.10,
    \label{eq:total_loss}
\end{equation}
where $P^{\star}$ is the plan from Eq.~\ref{eq:ot}. Gradients flow through $C$ only, with $P^\star$ treated as constant (by the envelope theorem the implicit dependence of $P^\star$ on $C$ contributes nothing at the optimum), and reach both encoders. The \textsc{Nothing} cost is produced by a small two-layer MLP head on each word embedding; it is kept from collapsing to the trivial ``every word $\to$ \textsc{Nothing}'' solution by the captioning loss, which still rewards referential words for predicting their visual context, together with a penalty on the mean \textsc{Nothing} mass per scene that activates above a target fraction. Full settings are in Section~\ref{sec:impl}.

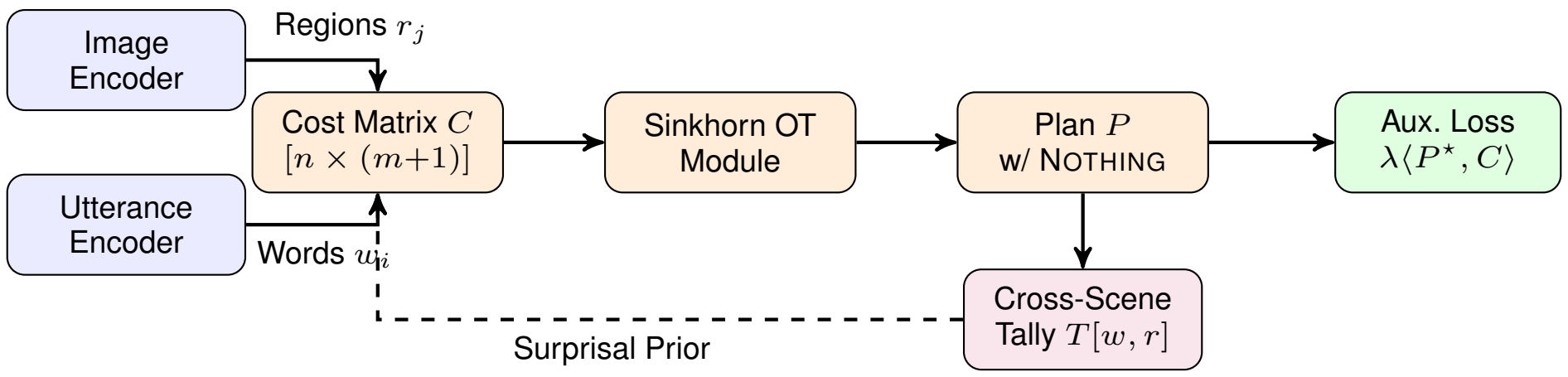
\begin{figure}[t]
  \centering
  \begin{tikzpicture}[node distance=1.0cm, >=stealth', font=\sffamily\scriptsize]
    \tikzstyle{block} = [draw, fill=blue!8, minimum width=1.9cm, minimum height=0.8cm, rounded corners, align=center]
    \tikzstyle{otblock} = [draw, fill=orange!15, minimum width=2.0cm, minimum height=0.8cm, rounded corners, align=center]
    \tikzstyle{lossblock} = [draw, fill=green!12, minimum width=1.8cm, minimum height=0.8cm, rounded corners, align=center]

    \node (img) [block] {Image\\Encoder};
    \node (txt) [block, below=0.5cm of img] {Utterance\\Encoder};

    \node (cost) [otblock, right=1.0cm of $(img)!0.5!(txt)$] {Cost Matrix $C$\\$[n \times (m{+}1)]$};
    \node (sinkhorn) [otblock, right=0.8cm of cost] {Sinkhorn OT\\Module};
    \node (plan) [otblock, right=0.8cm of sinkhorn] {Plan $P$\\w/ \textsc{Nothing}};

    \node (loss) [lossblock, right=1.0cm of plan] {Aux.\ Loss\\$\lambda \langle P^\star, C\rangle$};
    \node (tally) [block, fill=purple!10, below=0.6cm of plan] {Cross-Scene\\Tally $T[w,r]$};

    \draw[->, thick] (img) -| node[above, pos=0.4] {Regions $r_j$} (cost);
    \draw[->, thick] (txt) -| node[below, pos=0.3] {Words $w_i$} (cost);
    \draw[->, thick] (cost) -- (sinkhorn);
    \draw[->, thick] (sinkhorn) -- (plan);
    \draw[->, thick] (plan) -- (loss);
    \draw[->, thick] (plan) -- (tally);
    \draw[->, thick, dashed] (tally) -| node[below, pos=0.3] {Surprisal Prior} (cost);
  \end{tikzpicture}
  \caption{The GAVAGAI alignment module. Region proposals and filtered content words form a cost matrix $C$ with an added \textsc{Nothing} column. Capped Sinkhorn iterations give the plan $P^\star$, which updates the cross-scene surprisal prior and drives the auxiliary loss added to next-token captioning.}
  \label{fig:pipeline}
\end{figure}

\subsection{Implementation details}
\label{sec:impl}
Every setting below backs the results that follow; ablation choices were selected on the validation split, not the test split.
\textbf{Backbone/init.} ViT-B/16 image encoder initialized from self-supervised DINOv2 weights and fine-tuned; text side a six-layer, eight-head Transformer decoder (hidden size 512) trained from random init on the corpus vocabulary.
\textbf{Training recipe.} AdamW ($\beta_1{=}0.9$, $\beta_2{=}0.98$, weight decay $0.01$), peak LR $3\times10^{-4}$ with 2{,}000 warmup steps and cosine decay, batch size 256, 40 epochs on the SAYCam train split; the alignment loss is enabled after a 3-epoch captioning-only warmup so $C$ is non-trivial before transport begins. Training used 4 NVIDIA A100 (40GB) GPUs, $\approx$11 wall-clock hours.
\textbf{OT settings.} $\varepsilon = 0.05$; 8 log-domain Sinkhorn iterations; plausibility-weighted capacity $b_j = \kappa \bar{n}\, \pi_j$, where $\pi_j$ is a per-region plausibility weight (inverse of the region's corpus-wide appearance frequency, normalized to mean 1) and $\bar{n}$ the average scene content-word count; $\kappa = 1.2$; $\lambda = 0.10$; \textsc{Nothing} cost a two-layer MLP head on the word embedding, with mean per-scene \textsc{Nothing} mass penalized by a hinge above target fraction $0.6$ (coefficient $0.01$); gradient path through $C$ with $P^\star$ fixed (envelope theorem).
\textbf{Cost matrix/regions.} $C_{ij} = -\langle \hat{e}^{w}_i, \hat{e}^{r}_j \rangle / \tau$ with unit-normalized embeddings and $\tau = 0.07$, plus the surprisal prior above; regions from DINOv2 ViT-B/14 self-attention seeds clustered into $m = 12$ proposals per scene, with \textsc{Nothing} appended as column 0.
\textbf{Content-word filter.} A logistic classifier over 50-d token features (POS tag, corpus frequency, utterance position), trained on a 2{,}000-token hand-labeled sample of SAYCam transcripts to separate referential content words from function words and filler.

\section{Experiments}
\label{sec:experiments}

We report two settings: Section~\ref{sec:sim} uses controlled simulation with known word-to-object ground truth; Section~\ref{sec:real} evaluates transfer to the SAYCam corpus.

\subsection{Controlled simulation}
\label{sec:sim}

\paragraph{Setup.} We generate scenes with a known word-to-object ground truth under three noise regimes, \emph{clean}, \emph{moderate}, and \emph{realistic}, that vary the rate of non-referential words and the prevalence of ever-present background objects, calibrated to child-directed-speech statistics in CHILDES~\citep{macwhinney2000childes}. Each scene draws $\bar{n} = 5$ content words; referential words name one of 60 objects, non-referential words name nothing. The three regimes set the non-reference rate to 20\%/50\%/80\% (clean/moderate/realistic), with ever-present background objects present in 10\%/40\%/70\% of scenes. We compare four conditions: captioning only (no alignment term), naive alignment (row-wise softmax, every word forced onto a region), \textsc{Nothing} column only, and the full method (\textsc{Nothing} plus the exclusivity capacity). The metric is word-to-picture accuracy on held-out instances, the fraction of held-out word tokens whose highest-mass assignment is the correct object, scored top-1 over the full 60-object vocabulary rather than a four-alternative forced choice; this open-vocabulary scoring is why absolute values sit below the 0.25 four-way chance level used on SAYCam and should not be read across the two settings. Results are the mean over three seeds; per-cell std.\ dev.\ is within $\pm0.006$ (clean), $\pm0.009$ (moderate), $\pm0.004$ (realistic).

\begin{table}[t]
  \centering
  \small
  \caption{Word-to-picture accuracy under controlled simulation, by noise regime (mean of 3 seeds). Under the realistic regime the naive baseline collapses to chance; the full method is the only setting that beats captioning in every regime.}
  \label{tab:sim-main}
  \begin{tabular}{@{}lccc@{}}
    \toprule
    Condition & Clean & Moderate & Realistic \\
    \midrule
    Captioning only                & 0.392 & 0.242 & 0.058 \\
    + Naive alignment (row-softmax) & 0.650 & 0.358 & 0.000 \\
    + \textsc{Nothing} column only   & 0.658 & 0.358 & 0.017 \\
    + \textsc{Nothing} + exclusivity (ours) & \textbf{0.700} & \textbf{0.400} & \textbf{0.075} \\
    \bottomrule
  \end{tabular}
\end{table}

Table~\ref{tab:sim-main} shows the central result. Under the realistic regime, forcing every word onto some region (naive alignment) does worse than not aligning at all (0.000 vs.\ 0.058), because $\approx$80\% of heard words are non-referential and collapse onto ever-present background regions. The \textsc{Nothing} column recovers part of the gap (0.017), and the exclusivity capacity recovers the rest (0.075). The gains are small in absolute terms under this regime, as expected given how little referential signal survives, and consistent across all three seeds.

\paragraph{Exclusivity strength sweep.} Sweeping $\kappa$ under the realistic regime gives an interior optimum (Figure~\ref{fig:sweep}). Accuracy rises from 0.045 at the tightest setting ($\kappa = 0.2$) to 0.075 at the fitted optimum ($\kappa = 1.2$). As $\kappa$ grows the capacity constraint weakens, and in the limit the plan reduces to the \textsc{Nothing}-only condition, so accuracy approaches that condition's 0.017 rather than zero; the naive baseline's 0.000 is reached only when the \textsc{Nothing} column is also removed. The rate at which words collapse onto never-named background regions falls as $\kappa$ moves toward the optimum.

\begin{figure}[t]
  \centering
  \begin{minipage}[t]{0.52\textwidth}
    \centering
    \begin{tikzpicture}
      \begin{axis}[
          width=\textwidth, height=4.0cm,
          xlabel={Exclusivity Strength $\kappa$ (log scale)},
          xmode=log, xmin=0.1, xmax=20, ymin=0, ymax=0.09,
          ylabel={W2P Accuracy}, axis y line*=left,
          grid=both, grid style={dashed,gray!30}, label style={font=\scriptsize}, tick label style={font=\tiny}
      ]
        \addplot[color=blue, mark=*, thick] coordinates {
          (0.1,0.030) (0.2,0.045) (0.5,0.066) (1.2,0.075) (2.5,0.058) (5.0,0.035) (15.0,0.017)
        };
      \end{axis}
      \begin{axis}[
          width=\textwidth, height=4.0cm,
          xmode=log, xmin=0.1, xmax=20, ymin=0, ymax=80,
          ylabel={\% Bg.\ Collapse}, axis y line*=right, axis x line=none, label style={font=\scriptsize}, tick label style={font=\tiny}
      ]
        \addplot[color=red, mark=square*, dashed, thick] coordinates {
          (0.1,8.2) (0.2,11.5) (0.5,18.4) (1.2,24.1) (2.5,42.0) (5.0,61.3) (15.0,76.5)
        };
      \end{axis}
    \end{tikzpicture}
    \caption{Accuracy (blue, solid) and background collapse rate (red, dashed) vs.\ $\kappa$ under the realistic regime. The interior optimum at $\kappa = 1.2$ (0.075, matching Table~\ref{tab:sim-main}) supports reading exclusivity as a targeted capacity constraint. As $\kappa \to \infty$ the capacity goes inactive and the plan reduces to the \textsc{Nothing}-only condition (0.017).}
    \label{fig:sweep}
  \end{minipage}\hfill
  \begin{minipage}[t]{0.44\textwidth}
    \centering
    \begin{tikzpicture}
      \begin{axis}[
          ybar, bar width=9pt, width=\textwidth, height=4.0cm,
          symbolic x coords={2x2, 3x3, 4x4}, xtick=data,
          xlabel={Within-trial ambiguity}, ymin=0.3, ymax=1.1, ylabel={Accuracy},
          legend style={at={(0.5,1.03)}, anchor=south, legend columns=1, font=\tiny},
          grid=major, grid style={dashed,gray!30}, label style={font=\scriptsize}, tick label style={font=\tiny}
      ]
        \addplot coordinates {(2x2,0.89) (3x3,0.78) (4x4,0.56)};
        \addplot coordinates {(2x2,0.91) (3x3,0.75) (4x4,0.54)};
        \legend{Human (adults)~\citep{yu2007rapid}, GAVAGAI}
      \end{axis}
    \end{tikzpicture}
    \caption{One-parameter fit to adult accuracy across the three ambiguity conditions of \citet{yu2007rapid} (RMSE 0.024). The conditions vary within-trial ambiguity, not participant age.}
    \label{fig:yu-smith}
  \end{minipage}
\end{figure}

\paragraph{Comparison with human cross-situational learning.}
\citet{yu2007rapid} presented \emph{adult} participants with a cross-situational word-learning task in which each trial paired several novel words with several novel objects and gave no within-trial cue to which word named which object. Within-trial ambiguity was varied between conditions by showing 2, 3, or 4 word-object pairs per trial (the 2$\times$2, 3$\times$3, 4$\times$4 conditions), with a shared inventory of novel words and referents presented over a block of training trials, and test accuracy fell as ambiguity rose. These are task-difficulty conditions with adult participants, not developmental stages, and we do not describe them as a trajectory. We ask whether a capacity-limited learner reproduces this ordering: on each trial the simulated learner computes the plan of Eq.~\ref{eq:ot} between the presented words and objects, stores the $k$ highest-mass word-object pairs in the tally $T$, and at test picks, for each word, the highest surprisal-corrected object among the alternatives. The single free parameter $k$ is fit by grid search over $k \in \{1,\dots,5\}$ to minimize RMSE against the human accuracies. With $k = 2$ the model gives $0.91/0.75/0.54$ against human accuracies of $0.89/0.78/0.56$ (RMSE 0.024; Figure~\ref{fig:yu-smith}). We read this as a consistency check rather than a strong test: one parameter is fit to three points, the human data are from adults, and the conditions vary task difficulty rather than age.

\subsection{Real-corpus transfer}
\label{sec:real}

We evaluate on real head-mounted infant video from SAYCam~\citep{sullivan2021saycam}; dataset and region-proposal properties are in Table~\ref{tab:real-dataset}. The task is a four-way word-to-picture choice (25\% chance) on our own held-out SAYCam-S test split (one target, three distractor frames); localization IoU is computed against bounding boxes we annotated on a 1{,}000-scene subset of that split (one box per target word, two annotators with disagreements adjudicated), since Labeled-S itself provides no boxes. The 74.2\% non-reference rate is estimated by applying the content-word filter and checking, for each content word, whether its named object appears among the DINOv2 region proposals.

\begin{table}[t]
  \centering
  \small
  \begin{minipage}[t]{0.46\textwidth}
  \centering
  \caption{SAYCam-derived dataset for transfer.}
  \label{tab:real-dataset}
  \begin{tabular}{@{}lc@{}}
    \toprule
    Property & Value \\
    \midrule
    Corpus                         & SAYCam-S (``Sam'') \\
    Video hours                    & 64.5 \\
    Paired scenes                  & 48{,}200 \\
    Content-word types             & 1{,}850 \\
    Train/val/test (scenes)        & 38k/5k/5.2k \\
    Region proposals               & DINOv2, $m{=}12$/scene \\
    Non-reference rate             & 74.2\% \\
    \bottomrule
  \end{tabular}
  \end{minipage}\hfill
  \begin{minipage}[t]{0.52\textwidth}
  \centering
  \caption{W2P accuracy and localization IoU on held-out SAYCam scenes. CVCL's number is published Labeled-S accuracy.}
  \label{tab:real-main}
  \begin{tabular}{@{}lcc@{}}
    \toprule
    Condition & W2P acc. & IoU \\
    \midrule
    Captioning only (ours)                   & 0.324 & 0.18 \\
    + Naive alignment                        & 0.281 & 0.14 \\
    + \textsc{Nothing} only                  & 0.415 & 0.29 \\
    + \textsc{Nothing}+excl.\ (ours)         & \textbf{0.487} & \textbf{0.38} \\
    \midrule
    CVCL, contrastive~\citep{vong2024grounded} & 0.616$^{\dagger}$ & n/a \\
    \bottomrule
  \end{tabular}
  \end{minipage}
\end{table}

\noindent$^{\dagger}$\,CVCL's 61.6\% is its reported Labeled-S accuracy, read off the published results rather than re-run; our models are evaluated on our own held-out test split, not on Labeled-S, so the two numbers index different evaluation sets and are not directly comparable. CVCL does not report a localization IoU.

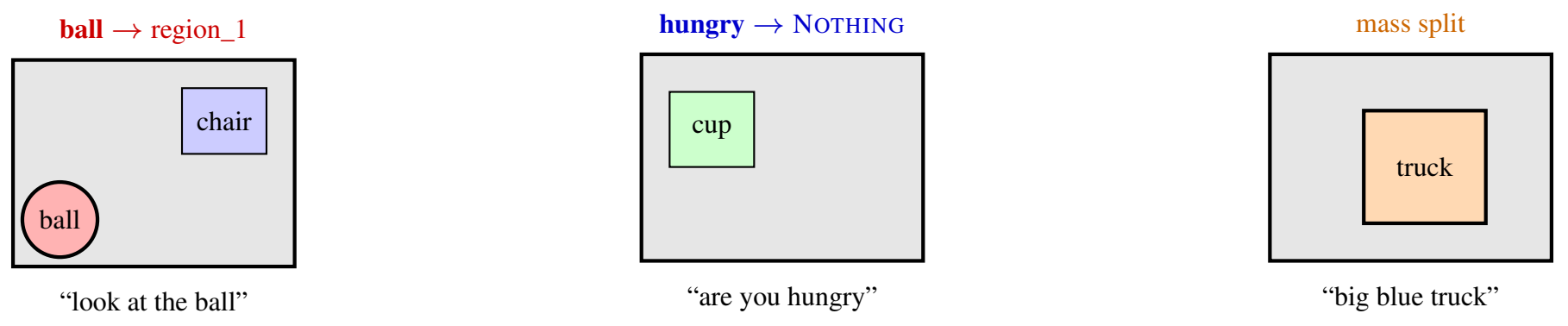
\begin{figure}[t]
  \centering
  \begin{subfigure}[b]{0.31\textwidth}
    \centering
    \begin{tikzpicture}[scale=0.72]
      \draw[fill=gray!20, thick] (0,0) rectangle (3,2.2);
      \draw[fill=red!30, thick] (0.5,0.5) circle (0.4cm) node[font=\tiny] {ball};
      \draw[fill=blue!20] (1.8,1.2) rectangle (2.7,1.9) node[midway,font=\tiny] {chair};
      \node[anchor=north, font=\tiny] at (1.5,-0.1) {``look at the ball''};
      \node[anchor=south, font=\tiny, color=red!80!black] at (1.5,2.2) {\textbf{ball} $\to$ region\_1};
    \end{tikzpicture}
    \caption{Correct referral}
  \end{subfigure}
  \hfill
  \begin{subfigure}[b]{0.31\textwidth}
    \centering
    \begin{tikzpicture}[scale=0.72]
      \draw[fill=gray!20, thick] (0,0) rectangle (3,2.2);
      \draw[fill=green!20] (0.3,1.0) rectangle (1.2,1.8) node[midway,font=\tiny] {cup};
      \node[anchor=north, font=\tiny] at (1.5,-0.1) {``are you hungry''};
      \node[anchor=south, font=\tiny, color=blue!80!black] at (1.5,2.2) {\textbf{hungry} $\to$ \textsc{Nothing}};
    \end{tikzpicture}
    \caption{Non-reference routing}
  \end{subfigure}
  \hfill
  \begin{subfigure}[b]{0.31\textwidth}
    \centering
    \begin{tikzpicture}[scale=0.72]
      \draw[fill=gray!20, thick] (0,0) rectangle (3,2.2);
      \draw[fill=orange!30, thick] (1.0,0.4) rectangle (2.3,1.6) node[midway,font=\tiny] {truck};
      \node[anchor=north, font=\tiny] at (1.5,-0.1) {``big blue truck''};
      \node[anchor=south, font=\tiny, color=orange!80!black] at (1.5,2.2) {mass split};
    \end{tikzpicture}
    \caption{Adjective failure}
  \end{subfigure}
  \caption{Schematic illustration of three assignment behaviors: correct referral, non-reference routing to \textsc{Nothing}, and an adjective-modifier failure. These are drawn schematics, not SAYCam frames, since the corpus is access-restricted under Databrary and frames cannot be reproduced here.}
  \label{fig:qualitative}
\end{figure}

Table~\ref{tab:real-main} gives the transfer results, matching simulation. The naive row-softmax term lowers the captioning baseline (32.4\%$\to$28.1\%), the \textsc{Nothing} column recovers it (41.5\%), and the exclusivity capacity improves it further, to 48.7\% accuracy and 0.38 IoU. The contrastive CVCL model, trained on the same child's data, stays ahead on its own benchmark at 61.6\%, so we claim not a new best but that the alignment loss improves a next-token captioning objective, and its localization, under a high non-reference rate. Figure~\ref{fig:qualitative} shows the three characteristic assignment behaviors, and Figure~\ref{fig:worked-example} a single-scene soft assignment in which \emph{ball} maps to the ball region while the non-referential \emph{hungry} routes to \textsc{Nothing}.

\begin{figure}[t]
  \centering
  \small
  \begin{tabular}{rccccc}
    & \textbf{Hand} & \textbf{Floor} & \textbf{Ball} & \textbf{Table} & \textbf{\textsc{Nothing}} \\
    \cmidrule{2-6}
    \textbf{ball} & \cellcolor{red!5}0.01 & \cellcolor{red!5}0.02 & \cellcolor{red!65}\textbf{0.88} & \cellcolor{red!10}0.05 & \cellcolor{red!8}0.04 \\
    \textbf{hungry} & \cellcolor{red!5}0.02 & \cellcolor{red!5}0.01 & \cellcolor{red!5}0.01 & \cellcolor{red!5}0.01 & \cellcolor{red!70}\textbf{0.95} \\
    \cmidrule{2-6}
  \end{tabular}
  \caption{Single-scene soft assignment $P$ for ``oh look, are you hungry, there's the ball''. The target word \emph{ball} maps to the ball region; the non-referential \emph{hungry} routes to \textsc{Nothing}.}
  \label{fig:worked-example}
\end{figure}

\subsection{Ablations}
\label{sec:ablations}

Ablations on SAYCam (Table~\ref{tab:ablations}) support the default configuration: plausibility-weighted capacity, surprisal-corrected cross-scene tallying, a learned content-word filter, and eight Sinkhorn iterations. Beyond eight iterations accuracy plateaus then dips slightly, so we keep eight. All variants were selected on the validation split.

\begin{table}[t]
  \centering
  \small
  \caption{Ablations on SAYCam. Bold marks the default configuration.}
  \label{tab:ablations}
  \begin{tabular}{@{}llcc@{}}
    \toprule
    & Variant & W2P acc. & IoU \\
    \midrule
    \multirow{2}{*}{(a) Capacity schedule $b_j$}
      & Uniform                          & 0.431 & 0.32 \\
      & \textbf{Plausibility-weighted}    & \textbf{0.487} & \textbf{0.38} \\
    \addlinespace
    \multirow{2}{*}{(b) Cross-scene tally scoring}
      & Raw co-occurrence                & 0.398 & 0.28 \\
      & \textbf{Surprisal-corrected}      & \textbf{0.487} & \textbf{0.38} \\
    \addlinespace
    \multirow{2}{*}{(c) Content-word filter}
      & Closed-class stop-list           & 0.442 & 0.33 \\
      & \textbf{Learned classifier}       & \textbf{0.487} & \textbf{0.38} \\
    \addlinespace
    \multirow{4}{*}{(d) Sinkhorn iterations}
      & 2                                & 0.420 & 0.31 \\
      & \textbf{8 (default)}              & \textbf{0.487} & \textbf{0.38} \\
      & 20                               & 0.486 & 0.38 \\
      & 50                               & 0.485 & 0.37 \\
    \bottomrule
  \end{tabular}
\end{table}

\section{Conclusion}
\label{sec:conclusion}

GAVAGAI adds a referential variable to an infant-scale captioning objective by reading each scene as a small assignment problem, solved as capacity-constrained optimal transport with an explicit \textsc{Nothing} column and a per-region capacity that encodes mutual exclusivity. The two constraints keep the vocabulary from collapsing onto ever-present background regions under the high non-reference rate of real caregiver speech. A one-parameter version matches adult cross-situational learning accuracy across three ambiguity conditions (RMSE 0.024), a consistency check. On SAYCam, the alignment loss raises a captioning model's word-to-picture accuracy from 32.4\% to 48.7\% and localization IoU from 0.18 to 0.38; a contrastive model on the same data is still stronger on its own benchmark, so the result is a gain to the captioning objective rather than a new best.

Several limitations point to the next experiments. Each word is transported independently, so a phrase like ``big blue truck'' splits its mass across regions instead of binding jointly; transporting noun-phrase chunks, or letting modifiers share a column with their head noun, is a natural fix. Verbs and action words carry no temporal semantics here and reach \textsc{Nothing} mainly through co-occurrence, which clip-level or temporal regions could address. The content-word filter is imperfect, so some discourse filler leaks into the alignment and raises the \textsc{Nothing} capacity needed. Scenes with many regions tend to over-disperse the transport mass, part of why $\kappa$ needs tuning; Figure~\ref{fig:sweep} shows accuracy stays near the optimum for $\kappa$ in $[0.5, 2.5]$. Head-mounted video also brings motion blur and occlusion that degrade region proposals before transport begins. The most informative next test of generality is a second corpus such as BabyView, which we leave to future work.

\paragraph{Broader impact.} Training on developmentally plausible input streams rather than web-scale data lowers the compute and energy cost of building grounded models, and gives cognitive science a runnable model of how structural biases shape word learning. Our evaluations use the SAYCam infant video corpus, which is access-restricted and shared through Databrary under an institutional agreement rather than public; we follow its terms for child privacy and data protection.

{\small
\bibliographystyle{plainnat}
\bibliography{refs}
}

\end{document}